\documentclass[conference]{IEEEtran}
\IEEEoverridecommandlockouts
\usepackage{cite}
\usepackage{amsmath,amssymb,amsfonts}
\usepackage{algorithmic}
\usepackage[pdftex]{graphicx}
\graphicspath{{./figures/}}
\usepackage{epsfig}
\usepackage{epstopdf}
\ifCLASSOPTIONcompsoc
    \usepackage[caption=false, font=normalsize, labelfont=sf, textfont=sf]{subfig}
\else
\usepackage[caption=false, font=footnotesize]{subfig}
\fi

\usepackage[T1]{fontenc}
\usepackage[utf8]{inputenc}

\makeatletter
\newcommand{\linebreakand}{%
  \end{@IEEEauthorhalign}
  \hfill\mbox{}\par
  \mbox{}\hfill\begin{@IEEEauthorhalign}
}
\makeatother

\usepackage{textcomp}
\usepackage{xcolor}
\def\BibTeX{{\rm B\kern-.05em{\sc i\kern-.025em b}\kern-.08em
    T\kern-.1667em\lower.7ex\hbox{E}\kern-.125emX}}

\makeatletter
\def\ps@IEEEtitlepagestyle{%
  \def\@oddfoot{\mycopyrightnotice}%
  \def\@evenfoot{}%
}
\def\mycopyrightnotice{%
  {\begin{minipage}{\textwidth}\footnotesize \copyright2019 IEEE.  Personal use of this material is permitted.  Permission from IEEE must be obtained for all other uses, in any current or future media, including reprinting/republishing this material for advertising or promotional purposes, creating new collective works, for resale or redistribution to servers or lists, or reuse of any copyrighted component of this work in other works.\end{minipage}}
  \gdef\mycopyrightnotice{}
}

\begin{document}

\title{Emotion Recognition System from Speech and Visual Information based on Convolutional Neural Networks}

\author{
\IEEEauthorblockN{Nicolae-Cătălin Ristea}
\IEEEauthorblockA{
\textit{University Politehnica of Bucharest}\\
Bucharest, Romania\\
r.catalin196@yahoo.ro}
\and
\IEEEauthorblockN{Liviu Cristian Duțu}
\IEEEauthorblockA{
\textit{Fotonation Romania, member of Xperi Group}\\
Bucharest, Romania\\
liviu.dutu@xperi.com}
\and
\IEEEauthorblockN{Anamaria Radoi}
\IEEEauthorblockA{
\textit{University Politehnica of Bucharest}\\
Bucharest, Romania\\
anamaria.radoi@upb.ro}
}


\IEEEoverridecommandlockouts
\IEEEpubid{\makebox[\columnwidth]{978-1-7281-0984-8/19/\$31.00~\copyright2019 IEEE.\hfill} \hspace{\columnsep}\makebox[\columnwidth]{ }}

\maketitle

\IEEEpubidadjcol

\begin{abstract}
Emotion recognition has become an important field of research in the human-computer interactions domain. The latest advancements in the field show that combining visual with audio information lead to better results if compared to the case of using a single source of information separately. From a visual point of view, a human emotion can be recognized by analyzing the facial expression of the person. More precisely, the human emotion can be described through a combination of several Facial Action Units. In this paper, we propose a system that is able to recognize emotions with a high accuracy rate and in real time, based on deep Convolutional Neural Networks.  In order to increase the accuracy of the recognition system, we analyze also the speech data and fuse the information coming from both sources, i.e., visual and audio. Experimental results show the effectiveness of the proposed scheme for emotion recognition and the importance of combining visual with audio data.  

\end{abstract}


\section{Introduction}
Facial attribute recognition, including facial action units and emotions, has been a topic of interest among computer vision researchers for over a decade. Being able to recognize and understand the emotion of a subject could be a key factor in a wide range of fields such as public security, healthcare (including therapeutic treatments) or entertainment. Moreover, the ability of today's systems to recognize and express emotions would leverage the barriers in obtaining a "natural" interaction between systems and humans. 

The interest of researchers on this subject lead to the development of Facial Action Coding System (FACS) \cite{tian2001recognizing}. Following FACS encoding system, each emotion can be modelled as a finite group of Facial Action Units (FAUs). Indeed, Ekman identifies several facial attributes that allow emotion recognition from face expressions (e.g., morphology, symmetry, duration, coordination of facial muscles) \cite{ekman1993facial}. 

The recent success of deep learning techniques in many Computer Vision-related applications influenced the emotion recognition field as well. Due to the release of labeled large datasets for emotion recognition from images, as well as the advances made in the design of convolutional neural networks, error rates have significantly dropped.  

An interesting approach towards emotion recognition is to use multimodal systems of recognition. It is well known that visual and audio information are very important when building emotion recognition systems because the usage of combined sound and video information leads to a better understanding of the emotion context than having access to a single source of information \cite{Marechal2019}. In this regard, several databases have been recently developed (e.g., CREMA-D \cite{cao2014crema}, OMG-Emotion \cite{barros2018omg}), but still the lack of multimodal data is a major problem. Therefore, despite the success of multimodal emotion recognition systems, there are still problems regarding emotion classification in real world scenarios that benefit from the existence of both visual and audio information.

In this paper, we propose a novel approach towards emotion recognition using multiple sources of information.  Our approach involves both visual and speech information and it is based on convolutional neural networks. The experiments prove the effectiveness of the proposed approach and show that having both images and sounds helps to achieve high classification accuracy.

In Section II, we review several of the state-of-the-art approaches in emotion recognition, whereas Section III describes several databases related to emotion recognition. The proposed approach is presented in Section IV and the corresponding results are discussed in Section V. Section VI concludes the paper.
 
\section{Related Work}

\subsection{Emotion recognition from images}

A wide range of approaches have been developed for the recognition of emotions from still images. The recognition system proposed in \cite{burkert2015dexpression} (called DeXpression) is based on a Convolutional Neural Network (CNN) architecture inspired by the well-known GoogleNet arhitecture \cite{GoogleNet}. The architecture contains two blocks of feature extraction composed by convolution, Max Pooling (MaxPool), Rectified Linear Units (ReLU) and concatenation layers. The end part of the network is a fully connected layer which performs the actual emotion classification. The authors use standard datasets (e.g. Extended Cohn-Kanade (CKP) and MMI Facial Expression Database) in their experimental setup and reported that the system has better performances than previous approaches. With an accuracy of 99.6\% for CKP and 98.63\% for MMI, the DeXpression architecture is robust, small and suitable for real-time applications when emotions are found in the pictures. However, the system fails to recognize an emotion from the first few frames, but these misclassification errors do not affect the overall accuracy of the system since the first corresponding instances can be considered as neutral. 

Another method, proposed by Z. Yu and C. Zhang \cite{yu2015image}, is based on learning multiple deep neural networks. The authors describe a more complex face detector composed by three state-of-the-art detectors, followed by a classification module made by combining multiple CNNs. The combining method considers minimizing both the log-likelihood loss and the hinge loss. The approach achieved state-of-the-art results on the Facial Expression Recognition (FER) Challenge 2013 dataset, whereas the classification accuracy reported on the validation and test set of Static Facial Expressions in the Wild (SFEW) dataset 2.0 was 55.96\% and 61.29\%, respectively.

A comprehensive review of the methods related to facial expression recognition can be found in \cite{DBLP:journals/corr/abs-1804-08348}. However, as the authors mention, two key issues appear when dealing with facial expression recognition system. Firstly, training deep convolutional neural networks require large volumes of annotated data. Secondly, variations such as illumination, head pose and person identity might lead to inconsistent recognition results. Bringing audio information to the recognition system may leverage several of these drawbacks.   

\subsection{Emotion recognition using visual and audio information}

Combining more sources of information leads to a higher accuracy rate if compared to the case of using a single source of information, audio or visual. EmoNets, proposed in \cite{kahou2016emonets}, is an emotion recognition system that considers visual features extracted using Convolutional Neural Networks, whereas a deep belief network is intended for building the representation of the audio stream. The spatio-temporal relations in the video streams are tackled by using a relational autoencoder. The authors stress that combining visual and audio features leads to a classifier that achieves a high accuracy rate. The method proposed in \cite{kahou2016emonets} won the 2013 Emotion Recognition in the Wild Challenge (EmotiW) \cite{dhall2013emotion}, reporting an accuracy level on the test set of 47.67\% for the 2014 dataset.

A different approach is presented in \cite{8070966} which considers using a CNN to extract features from the speech, whilst, in order to represent the visual information, a deep residual network ResNet-50 is used. The features from the before-mentioned networks are concatenated and inserted into a two-layers Long Short-Term Memory (LSTM) module. Two continuous values are predicted at each moment, namely arousal and valence. The method outperforms other previous methods on the RECOLA database of the Audio-Visual Emotion Challenge (AVEC) 2016.

\begin{figure}[!t]
  \centering
  \includegraphics[width=0.47\textwidth]{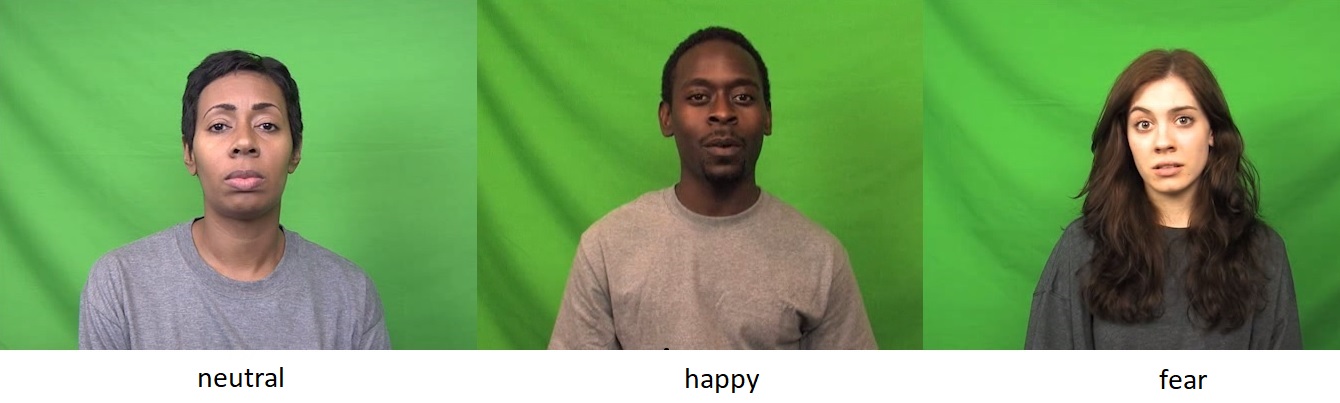}
  \caption[]{Examples of frames from the CREMA-D dataset along with the corresponding emotion labels.}
  \label{fig:crema-d}
\end{figure} 

\section{Database}

Several databases have been developed for emotion recognition purposes. However, regarding multimodal emotion recognition, the lack of data is a major problem. In this approach we focused on the CREMA-D database which is commonly used in multimodal emotion recognition \cite{cao2014crema, Vougioukas2019RealisticSF, beard-etal-2018-multi}. Some examples are provided in Fig.~\ref{fig:crema-d}.

The CREMA-D database was published in 2015 and contains 7442 clips of 91 actors (48 male and 43 female) with different ethnic backgrounds, coordinated by professional theater directors. The actors were asked to convey particular emotions while producing, with different intonations, 12 particular sentences that evoke the target emotions. There were six labeled emotions (neutral, happy, anger, disgust, fear, sad) and four different emotion levels (low, medium, high, unspecified). 

The recordings were performed in three different modes, namely audio, visual, and audiovisual. The labels corresponding to each recording were collected using crowdsourcing. More precisely, 2443 participants were asked to label the perceived emotion and its intensity in three modes, depending on the information put at their disposal: video, audio and combined audiovisual. The human accuracy reported for each mode is presented in Table~\ref{table: human accuracy CREMA-D}. In this case, human training was achieved through participants' previous experiences, whereas the results mentioned in Table~\ref{table: human accuracy CREMA-D} refer to the accuracy over the whole dataset.

\begin{table}[t]
\caption{Human accuracy rate on CREMA-D \cite{cao2014crema}} 
\label{table: human accuracy CREMA-D}
\begin{center}
\begin{tabular}{ |c | c | c | c|} 
\hline
\textbf{Mode} & Audio & Video & Audio + Video \\ 
\hline
\textbf{Accuracy} & 40.9\% & 58.2\% & 63.6\% \\ 
\hline
\end{tabular}
\end{center}
\end{table}

\begin{figure*}[!t]
  \centering
  \includegraphics[width=1\textwidth]{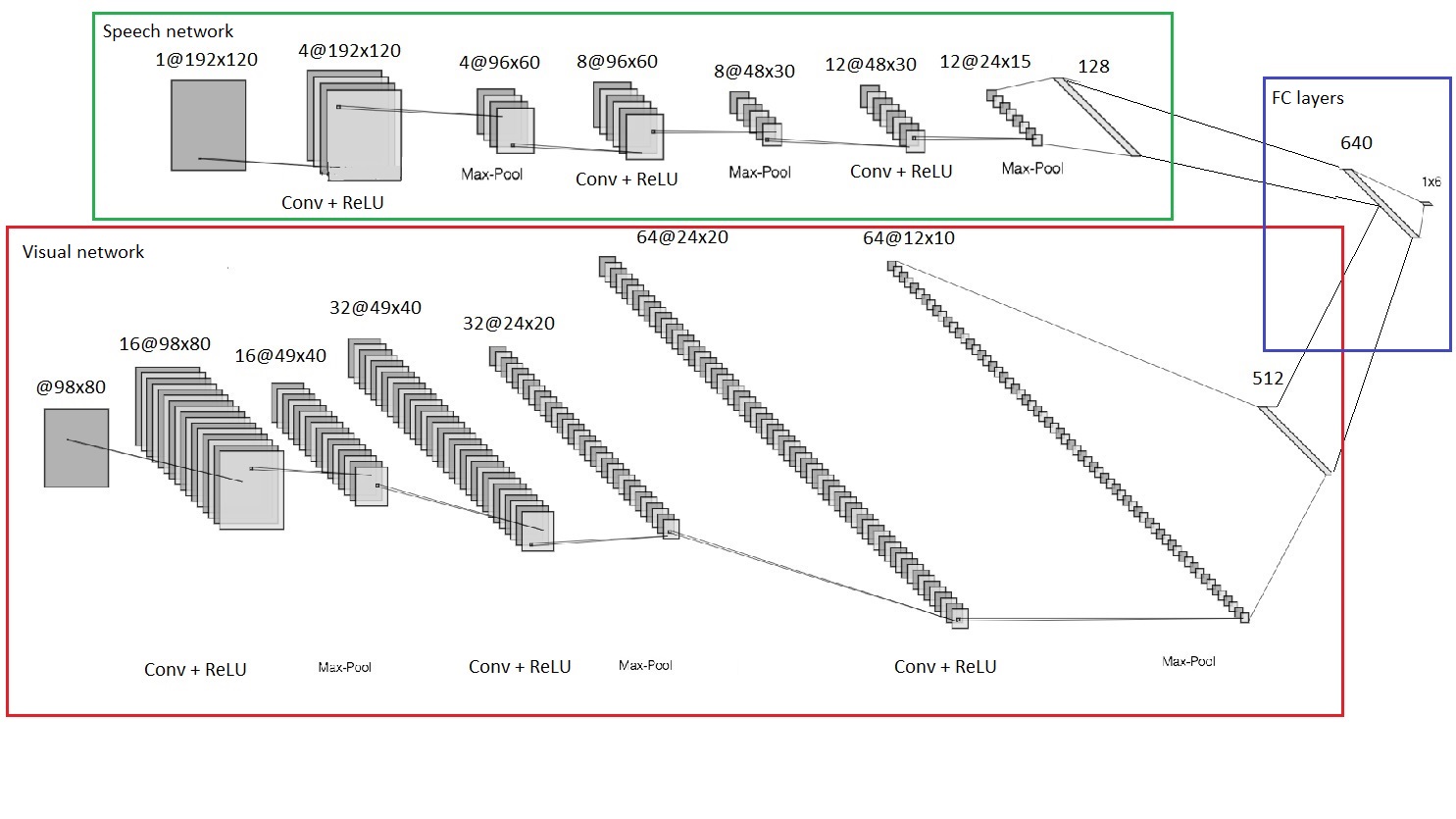}
  \caption[]{Proposed scheme for emotion recognition. The first part of the network deals with visual information extraction, whilst, the second part of the network handles the sound information. The last part of the network is a stack of two fully-connected (FC) layers, representing the classifier.}
  \label{fig:ar5}
\end{figure*}

\section{Proposed method}

\subsection{Spectrogram}
The \textit{spectrogram} represents a traditional approach towards the time-frequency analysis of signals with many applications in the fields of music, radar and speech processing. More precisely, the spectrogram of a signal shows the temporal evolution of its short-time spectral amplitude. From a visual representation point of view, the spectrogram is a two-dimensional plot with \textit{x} axis representing time, \textit{y} axis representing frequency and the magnitude of signal being encoded by the pixel value.

The most common method of computing a spectrogram is with the discrete Short-Time Fourier Transform (STFT), described by the following formula:

\begin{equation}
    STFT\{x[n]\}(m, k) = \sum_{m=-\infty}^{\infty} x[m] \cdot w[n-m] e^{-j \frac{2 \pi}{N_x}k n}
\end{equation}
where $N_x$ is the number of samples.

Moreover, several others methods of computing a spectrogram have developed during time. The \textit{Wigner-Ville} distribution is one of the techniques used to obtain information about a signal which varies in time \cite{cohen1989time}. Wavelet analysis, through the \emph{continuous wavelet transform} with different wavelet bases can also be used for this purpose. Nevertheless, in this study we restrict ourselves to the Short-Time Fourier Transform as a way to compute the spectrogram, and leave the other methods presented above for future investigations.

While all these techniques extract a time-frequency representation of the signal's energy density, we note that the resolution accuracy of all these representations is bounded by the time-frequency uncertainty principle (see \cite{cohen1995time}, Chapter 3) which forbids arbitrarily accurate energy density representations in both time and frequency simultaneously. This is why we believe that in order to extract relevant features for the emotions classification task, these time-frequency representations should be followed by a learned hierarchical feature extractor, such as a convolutional neural network, as detailed below.

\subsection{Preprocessing}
The video sequence is divided into frames and we consider keeping a fixed number of images $N$, that are equally distanced in the video. The following step consists in detecting the face of the subjects and resizing them to have equal size. In order to prevent the overfitting effect caused by the insufficient number of training samples, data augmentation techniques are applied, e.g. rotation, random cropping, different illumination and flips. In this sense, for each video sequence in part, other 30 videos were made following the methods presented above.

\begin{figure}[t]
     \centering
     \subfloat[Angry]{
         \includegraphics[width = 0.45\textwidth]{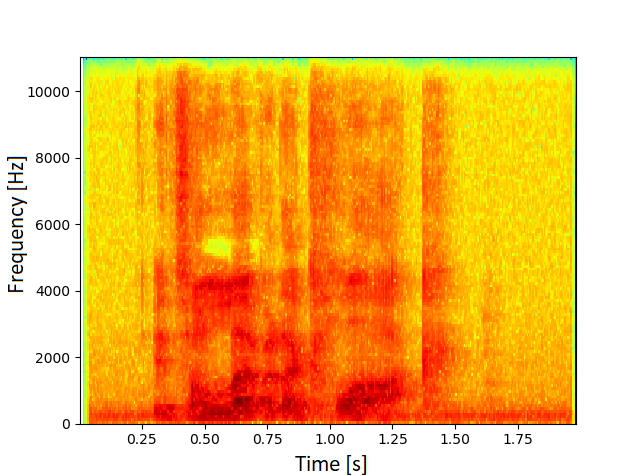}
     }
     \hfill
     \subfloat[Happy]{
         \includegraphics[width = 0.45\textwidth]{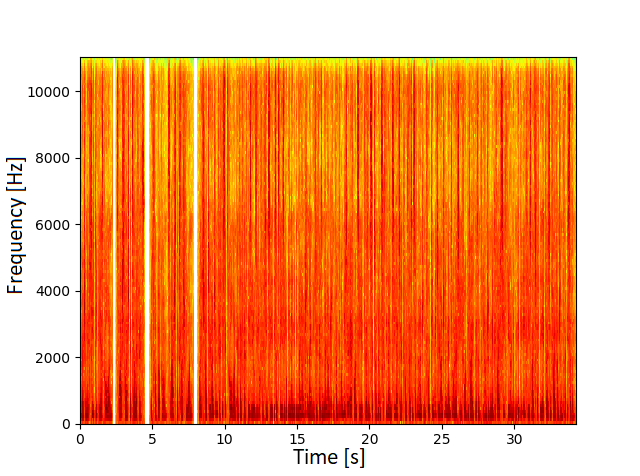}
     }
        \caption{Resized spectrograms for two different types of emotions}
        \label{fig: spectograms}
\end{figure}

The raw audio signal is processed with the discrete Short-Time Fourier Transform (STFT) described above.
In the majority of cases, the audio files do not have the same length. For this reason, the spectrograms do not have the same width. In order to overcome this aspect, the spectrograms are reshaped to a fix dimension (the height is the same for all signals, so a reshape was made on the $x$ axis). Two examples of spectrograms for two audio signals expressing different types of emotions are shown in Fig.~\ref{fig: spectograms}.

\begin{figure*}[!t]
  \centering
  \includegraphics[width=0.9\textwidth]{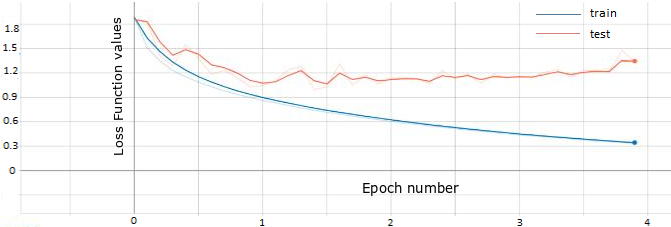}
  \caption[]{Loss function scores on the training set and sets.}
  \label{fig:loss function}
\end{figure*}

\begin{figure}[!t]
  \centering
  \includegraphics[width = 0.49\textwidth]{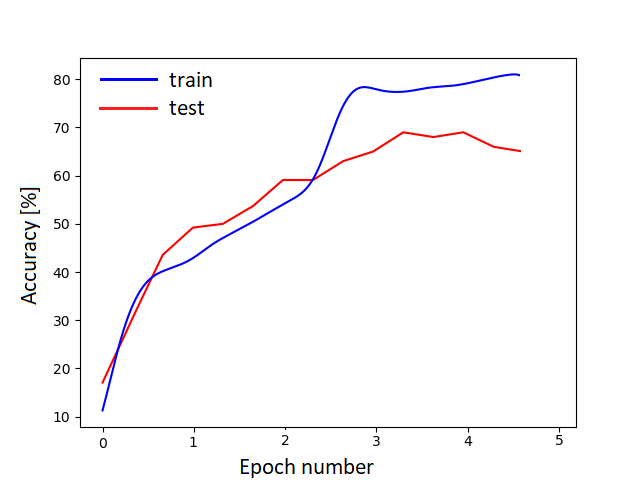}
  \caption[]{Accuracy scores on the training set and sets.}
  \label{fig: accuracy}
\end{figure}

\subsection{Proposed Architecture}
In this paper, we propose a deep Convolutional Neural Network (CNN)-based architecture, composed of three parts. The first part of the network deals with the feature extraction from the image sequences, the second part with the feature extraction from the audio signals, whereas the last part performs the emotion recognition part. The entire architecture is depicted in Fig.~\ref{fig:ar5}.

An important component of the proposed network architecture is the convolution operation, which is defined, for the audio (1-D) and visual (2-D) signals as follows:

\begin{equation}
    (f  * h)[i] = \sum_{k=-T}^T h[k] \cdot f[i-k]
\end{equation}

\begin{equation}
    (f  * h)[i, j] = \sum_{k=-T}^T \sum_{m=-T}^T h[k,m] \cdot f[i-k,j-m]
\end{equation}
where $h[k]$ and $h[k,m]$ are the 1-D and 2-D kernels whose parameters are learned during the training phase, and $f$ is the 1-D or 2-D signal at hand.

The first part of the network is represented by the lower branch of the scheme presented in Fig.~\ref{fig:ar5} and it is designed to receive a sequence of images and to output abstract features which encode the information from frames. The input dimension is $B \times N \times W \times H$, where $B$ is the batch size, $N$ is the length of the video sequence, whereas $W$ is the width and $H$ is the corresponding height. The video frames are analysed one after another.  

 The second part of the network is also a CNN-based architecture whose aim is to process the spectrograms of the audio signals and to extract meaningful features from them. Two remarks should be made at this point. Firstly, the number of convolutional layers is rather small. Secondly, the last layer contains a smaller number of neurons if compared to the first part of the network which deals with image analysis. This is argued by the fact that images contain more information than the audio signals (e.g., the position of the eyebrows, eyes, mouth or cheeks may indicate person's emotional state). More specifically, non-verbal information is essential in discovering the emotional state of an individual. For example, happiness can be unambiguously determined from the facial expression of a person. However, vocal expression is needed to detect anger \cite{cao2014crema}.  

 The third part of the network shown in Fig.~\ref{fig:ar5} is a classifier composed of two fully-connected (FC) layers.  After the inference through the before-mentioned parts of the network, the abstract features are concatenated and the resulting feature vector is introduced in the last part of the network, which provides the probability distribution of the emotion in the analyzed video. As already mentioned,  the lengths of the feature vectors for frames and sound are different, in proportion of 4:1, because the importance of the visual data is greater.
 
 The output is a tensor of $L$ elements $\{y_{x,1}, y_{x,2}, \ldots, y_{x,L} \}$ that indicates the emotional state of the person (i.e., $L$ being the number of emotional states taken into consideration). In order to obtain the probability distribution for the emotional states, we consider using a SoftMax function that transforms the vector of 6 scores $\{y_{x,1}, y_{x,2}, \ldots, y_{x,L} \}$  into a normalized vector of probabilities $\{p_{x,1}, p_{x,2}, \ldots, p_{x,L} \}$, using the formula:
 \begin{equation}
     p_{x,c} = \frac{e^{y_{x,c}}}{\sum_{c'=1}^{L} e^{y_{x,c'}}}
 \end{equation}
 for $c \in \{1,..., L\}$.

The loss function considered for training is the Cross-Entropy, defined, for an observation $x$, as:
\begin{equation}
    \mathcal{L}(x) = -\sum_{c=1}^L \delta_{x,c} \log (p_{x,c})
\end{equation}
where $\delta_{x,c}$ is the binary indicator (0 or 1) showing if class label $c$ is correct for observation $x$ or not and $p_{x,c}$ is the predicted probability of observation $x$ pertaining to class $c$. 
  
\section{Experimental Setup and Results}

In order to prove the effectiveness of the proposed classification architecture that the audio data is helpful for the classification scope, we did two experiments. In the first experiment, we consider as input data just the sequence of images and we removed the part of the network which processes spectrograms. In the second experiment, we used the entire network architecture described in the previous section which receive data from two different sources, i.e., audio and visual. 
Several setups are considered during the implementation of the proposed neural network. As mentioned above, the spectrograms of the signals must have equal dimensions even if the analysed audio signals are different in length. We choose to resize all the spectrograms to $192 \times 120$. Similarly, the images containing the detected faces are resized to images of $98 \times 80$ pixels.

The experiments were implemented in PyTorch on a computer with NVIDIA GTX 1080Ti graphic board and Intel i7 processor. We tried more optimizers, but the best performance regarding time of convergence and network accuracy was achieved by Adam. The learning rate was $10^{-4}$, with a weight decay of $5 \cdot 10^{-5}$.

At the training phase, we set the batch size to be 32 and we kept 20 frames equally distanced from every video. Therefore, the input data size is $32 \times 20 \times 98 \times 80$ for the first part of the network and $1 \times 1 \times 192 \times 120$ for the second part of the network. In order to learn the parameters that minimize the loss function, we have used a learning rate of $0.001$ and Adam as the method for stochastic optimization \cite{Adam}.

The testing method is \textit{leave-one-out}. More precisely, we trained the network on all actors except one, then we tested the network on the actor that was removed from the dataset. This procedure was made for all 91 actors from database and the obtained results were averaged to obtain the final classification accuracy score.

The loss function values obtained on train and test sets are showed for different epoch numbers in Fig.~\ref{fig:loss function}, whereas the corresponding accuracy values are given in Fig.~\ref{fig: accuracy}. It can be easily observed that the loss function, computed for the training set, has a fast decay in the first epoch, followed by a slower decay in the next epochs, as showed in Fig.~\ref{fig:loss function}.

We mention that the network training was stopped after four epochs in average, just before overfitting occurs. Moreover, as shown in Fig.~\ref{fig: accuracy}, the accuracy starts to decay at the fifth epoch. Also, the validation loss was greater in some cases because the Cross-Entropy loss function is known to punish significantly the network if the classification returns unsuitable results. Because the overfitting was a constant problem in the training process, new methods of data augmentation could be added, such as near infra–red (NIR) generated data \cite{mualuaescu2019improving}.

The obtained results on the test set are reported in Table~\ref{table: accuracy} for two different modes of the recordings, namely video and audiovisual, respectively. It is worth mentioning that the accuracy is reported considering the leave-one-out strategy, whereas in Table~\ref{table: human accuracy CREMA-D} the human accuracy was computed over the whole dataset. 

\begin{table}[t]
\caption{Classification results on CREMA-D} 
\begin{center}
\begin{tabular}{ |c | c | c | c | c|} 
\hline
 & Mean accuracy over 91 actors & Standard deviation \\ 
\hline
\textbf{Video} & 62.84\% & 14.06 \\
\hline
\textbf{Audio + Video} & 69.42\% & 13.25 \\
\hline
\end{tabular}
\end{center}
\label{table: accuracy}
\end{table}

\section{Conclusion}
In this article, we present a network architecture that is able to recognize emotion by combining visual with audio information. The output of the network is a distribution of probabilities for each type of emotion considered for training.

The architecture is CNN-based, whilst the time series of frames are processed in a pipeline that takes into consideration multiple channels for each image. This approach introduced the data variation in time, which is a key point in emotion recognition because each emotion evolves in intensity over time.

The results sustain the idea that having more information about a subject is important for determining the emotion with increased accuracy. In this sense, it can be observed that adding the audio information (i.e., which is a signature for each subject) in the network, yield an improvement of almost 6.57\% in the accuracy results.

Also, we can observe that having multimodal information about a subject the standard deviation of predictions is lower, which sustains the fact that the network performs better when the input is a more complex information, in our case video and audio information.

Moreover, our results, in both experiments, outperform the human accuracy reported by the authors of CREMA-D dataset, in video mode with 4.64\% and, in video combined with audio, with 5.82\%.

As future development, the emotion intensity levels reported on CREMA-D samples could be taken into consideration to further increase the accuracy of the emotion recognition system. This information may help the training process by pondering the relative importance of every sample. Other future research directions include the use different techniques for time-frequency analysis, such as the Wigner-Ville distribution or the Continuous Wavelet Transform, which could lead to improved information extraction from the audio signal.  



\section*{Acknowledgment}
This work has been partially supported by the Ministry of Innovation and Research, UEFISCDI, project SPIA-VA, agreement 2SOL/2017, grant PN-III-P2-2.1-SOL-2016-02-0002.

\bibliographystyle{IEEEtran}
\bibliography{references}

\end{document}